\documentclass[10pt,conference,letterpaper]{IEEEtran}
\IEEEoverridecommandlockouts

\usepackage[
  letterpaper,
  top=1.03in,
  bottom=0.92in,
  left=0.82in,
  right=0.82in,
  columnsep=0.20in
]{geometry}

\renewcommand{\baselinestretch}{0.95}
\usepackage{cite}
\usepackage{amsmath,amssymb,amsfonts}
\usepackage{algorithmic}
\usepackage[ruled,vlined]{algorithm2e}
\usepackage{graphicx}
\graphicspath{{./Figures/}}
\usepackage{textcomp}
\usepackage{xcolor}
\usepackage{booktabs}
\usepackage{threeparttable}
\usepackage{multirow}
\usepackage{svg}
\usepackage[hidelinks]{hyperref}

\usepackage[compact]{titlesec}
\titlespacing{\section}{0pt}{0.9ex plus 0.2ex minus 0.1ex}{0.5ex plus 0.1ex}
\titlespacing{\subsection}{0pt}{0.8ex plus 0.2ex minus 0.1ex}{0.4ex plus 0.1ex}
\titlespacing{\subsubsection}{0pt}{0.6ex plus 0.1ex minus 0.1ex}{0.3ex plus 0.1ex}
\titlespacing{\paragraph}{0pt}{0.5ex plus 0.1ex minus 0.1ex}{0.6em}

\makeatletter
\def\@oddfoot{\hfil\thepage\hfil}
\def\@evenfoot{\hfil\thepage\hfil}
\makeatother

\def\BibTeX{{\rm B\kern-.05em{\sc i\kern-.025em b}\kern-.08em
    T\kern-.1667em\lower.7ex\hbox{E}\kern-.125emX}}

\begin{document}

\title{Towards Long-Horizon Vessel Trajectory and Destination Forecasting with Reasoning Large Language Models}

\author{
	\parbox{\textwidth}{%
		\centering
		Hongwei Wang$^{1}$, Miao Zhou$^{2}$, Fengde Wang$^{2}$, Yuting Wang$^{2}$,\\
		Jiewen Yu$^{2}$, Jun-Yan He$^{3}$, Bohao Qu$^{4}$, Wanbing Zhang$^{1}$,\\
		Xiuju Fu$^{1}$, Qing Guo$^{5}$, Zipei Fan$^{6}$, Yingying Xing$^{2*}$, Yi Yuan$^{6*}$%
	}%
    \thanks{\scriptsize $^{1}$~Institute of High Performance Computing (IHPC), A*STAR, Singapore 138632, Singapore. Emails: Hongwei Wang (\mbox{wang\_hongwei@a-star.edu.sg}), Wanbing Zhang (\mbox{zhangwab@a-star.edu.sg}), Xiuju Fu (\mbox{fuxj@a-star.edu.sg}).}%
	\thanks{\scriptsize $^{2}$~The Key Laboratory of Road and Traffic Engineering, Ministry of Education, Tongji University, Shanghai 201804, China. Emails: Miao Zhou (\mbox{2253932@tongji.edu.cn}), Fengde Wang (\mbox{2254342@tongji.edu.cn}), Yuting Wang (\mbox{2051920@tongji.edu.cn}), Jiewen Yu (\mbox{2352661@tongji.edu.cn}), Yingying Xing (\mbox{yingying199004@tongji.edu.cn}).}%
	\thanks{\scriptsize $^{3}$~Meituan Inc., Shenzhen, China. Email: Jun-Yan He (\mbox{junyanhe1989@gmail.com}).}%
	\thanks{\scriptsize $^{4}$~Centre for Frontier AI Research (CFAR), A*STAR, Singapore 138632, Singapore. Email: Bohao Qu (\mbox{qubohao@ieee.org}).}%
	\thanks{\scriptsize $^{5}$~Nankai University, Tianjin 300071, China. Email: Qing Guo (\mbox{tsingqguo@ieee.org}).}%
	\thanks{\scriptsize $^{6}$~School of Artificial Intelligence, Jilin University, Changchun, Jilin 130012, China. Emails: Zipei Fan (\mbox{fanzipei@jlu.edu.cn}), Yi Yuan (\mbox{yuanyi23@mails.jlu.edu.cn}).}%
	\thanks{\scriptsize $^{*}$~Corresponding authors: Yingying Xing, Yi Yuan.}%
}

\maketitle

\pagestyle{plain}
\pagenumbering{arabic}

\begin{abstract}

Long-horizon maritime trajectory prediction is important for shipping management, logistics planning, and maritime risk analysis, yet month-level forecasting remains insufficiently studied. Existing deep learning methods mainly focus on short- and mid-term coordinate extrapolation and often struggle to preserve route feasibility and destination correctness over extended horizons. 
This paper investigates joint long-horizon vessel trajectory and destination forecasting with reasoning-capable large language models, and develops a Maritime LLM post-training framework based on Reinforcement Learning with Verifiable Reward (RLVR).
An AIS-based benchmark is constructed with 60-day historical trajectories and 30-day forecasting horizons, in which vessel trajectories are converted into semantic textual representations for RL prompt construction. The RLVR framework aligns LLMs with maritime forecasting objectives by enforcing physical validity, providing early-weighted process-level trajectory supervision, and evaluating destination correctness through hierarchical matching and curriculum learning. Experimental results show that RLVR-trained LLMs substantially improve over zero-shot LLMs and widely used deep learning baselines, especially on destination-related metrics. 
Among the evaluated RLVR-trained variants, 4B LLMs achieve the best overall performance, suggesting that reward-compatible optimization and task-specific capacity matching are more important than simply using larger 8B or 14B LLMs under the current structured forecasting setting. 
The deep learning comparison further shows that LSTM remains a strong deep learning baseline under the current LLM post-training setting with limited fine-tuning data, while Transformer-style spatio-temporal models typically require larger datasets and richer structured inputs. 
These results indicate that the RLVR-trained LLM better exploits semantic voyage context for long-horizon maritime forecasting. Overall, this work provides a practical step toward semantic, verifier-aligned maritime forecasting for shipping operations and maritime decision support.

\end{abstract}


\section{Introduction}
\label{sec:introduction}

Maritime trajectory prediction is important for shipping management, logistics planning, and maritime situational awareness. While existing studies have achieved strong results on short-term forecasting, most existing models focus on local coordinate evolution over relatively short horizons. In contrast, long-horizon prediction over weeks or months is more relevant for strategic tasks such as berth allocation, fleet scheduling, resource planning, and risk mitigation, because long-horizon prediction reflects destination trends, route-level behavior, and voyage intent. Despite its practical importance, month-level maritime forecasting remains largely underexplored.

\begin{figure}[t]
\centering
\includegraphics[width=\columnwidth]{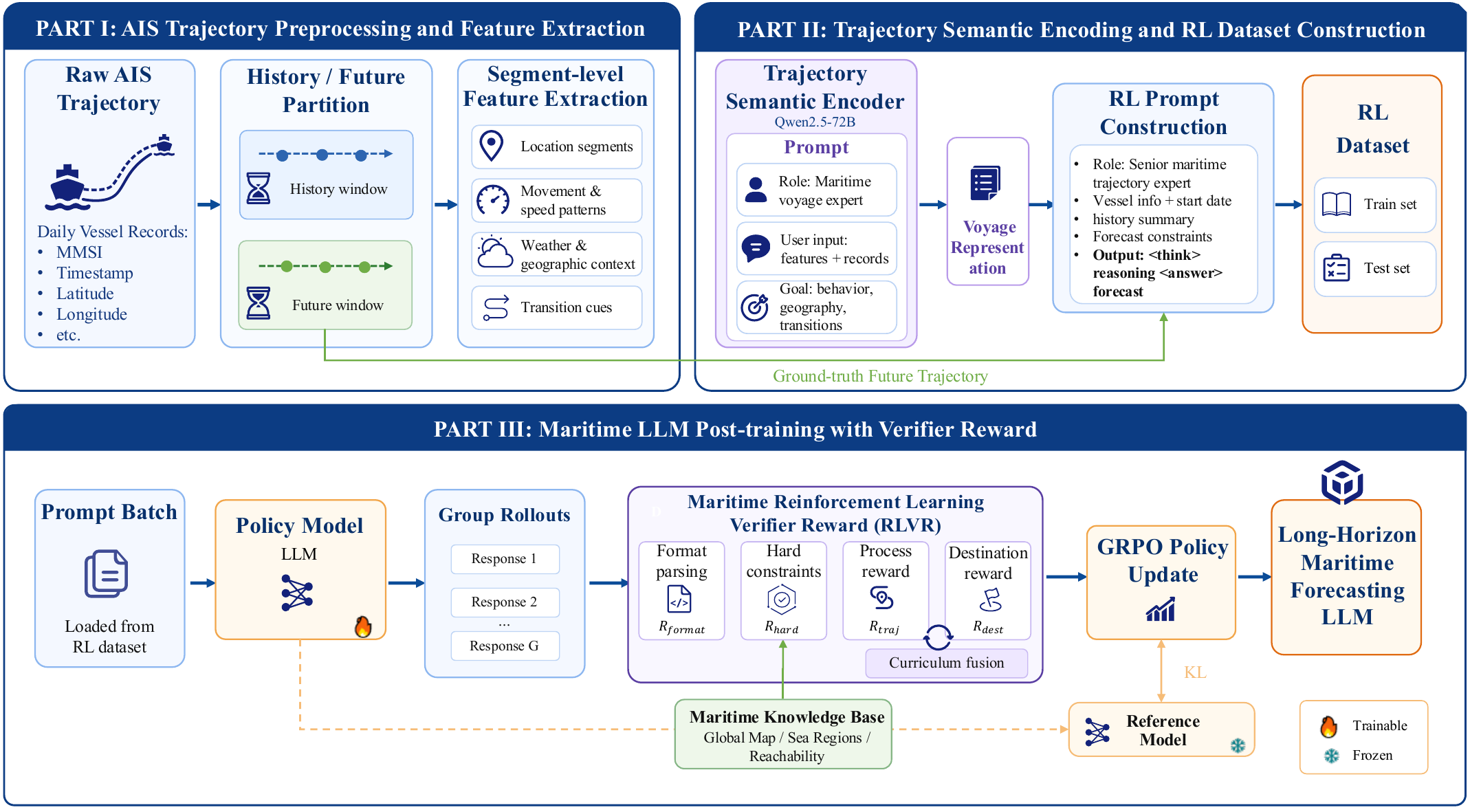}
\caption{Overall framework of the proposed model. AIS trajectories are encoded into semantic textual representations for RL prompt construction, and the policy LLM is post-trained with a maritime verifiable reward.}
\label{fig:framework}
\end{figure}


A core challenge is that long-horizon vessel movement is not determined solely by local motion dynamics, but is also shaped by destination planning, route feasibility, maritime connectivity, weather, and operational constraints. As a result, conventional deep learning models trained with point-wise regression losses often struggle to maintain route consistency and destination correctness when the prediction horizon extends to the month level.

Large language models (LLMs) provide a promising alternative. The long-context modeling and reasoning capabilities of LLMs make these models better suited for capturing high-level voyage semantics and structured decision patterns beyond coordinate extrapolation. Motivated by these challenges, we propose a reasoning-oriented LLM framework for joint long-horizon vessel trajectory and destination forecasting.

The contributions of this paper are as follows:
\begin{enumerate}
    \item We formulate a \textbf{joint long-horizon vessel trajectory and destination forecasting} task based on 60-day AIS history and 30-day future prediction, and develop an LLM-oriented data construction pipeline with trajectory preprocessing, semantic encoding, and RL prompt construction.
    \item We introduce a \textbf{Maritime Reinforcement Learning with Verifiable Reward (RLVR)} framework, where the reward combines hard-constraint gating, early-weighted process reward, and hierarchical destination matching with curriculum learning.
    \item We train a \textbf{long-horizon maritime forecasting LLM} with verifier-guided GRPO and show that the RLVR-trained model outperforms zero-shot LLMs and widely used deep learning baselines, especially on destination-related metrics.
\end{enumerate}

\section{Related Work}
\label{sec:relatedwork}

\subsection{Deep Learning for Maritime Trajectory Prediction}

Deep learning has substantially advanced maritime trajectory prediction, especially in short-term settings. Early recurrent models such as LSTM~\cite{hochreiter1997long} and GRU~\cite{el2023deep} established strong sequence modeling baselines, while attention-based and Transformer architectures further improved temporal representation learning~\cite{vaswani2017attention,wang2024advancements}. In time-series forecasting, specialized models such as Informer~\cite{zhou2021informer}, Autoformer~\cite{wu2021autoformer}, and PatchTST~\cite{nie2023patchtst} improved long-sequence efficiency and representation power.

These methods have shown strong performance on short- and mid-term prediction, but the formulation remains largely point-wise and regression-driven. For month-level maritime forecasting, such a point-wise formulation is limiting because long-horizon vessel behavior depends on recent coordinates as well as route feasibility, destination intent, and broader maritime structure. Existing deep learning models therefore struggle to preserve route-level consistency and destination correctness over extended horizons.

\subsection{LLM-based Spatio-Temporal Modeling}

Recent studies have explored large language models for spatio-temporal prediction, leveraging the reasoning ability and contextual knowledge integration capacity of LLMs. One line of work adapts structured spatio-temporal inputs to LLMs through tokenization or embedding schemes, including ST-LLM~\cite{liu2024spatial}, Time-LLM~\cite{jin2023time}, and UrbanGPT~\cite{li2024urbangpt}. Another line frames forecasting as an explicit reasoning problem, in which language models infer future behavior through structured prompts and reward-guided optimization. Time-R1~\cite{luo2025time}, for example, showed that reinforcement-based fine-tuning can improve reasoning ability in time-series forecasting.

In maritime applications, early LLM-based efforts such as AIS-LLM~\cite{park2025ais} have begun exploring joint prediction and anomaly detection, but still focus mainly on short-term settings. In contrast, month-level maritime forecasting requires reasoning over voyage intent, destination planning, and route-level feasibility. These challenges motivate our verifier-guided RL framework, which explicitly combines physical validity, process-level trajectory supervision, and hierarchical destination optimization for long-horizon maritime forecasting.

\section{Methodology}
\label{sec:methodology}

\subsection{Problem Formulation}

We consider long-horizon maritime trajectory and destination forecasting. Given a vessel's 60-day historical trajectory
\begin{equation}
\mathbf{X}=\{x_t\}_{t=1}^{60}, \qquad
x_t=(d_t,\mathrm{lon}_t,\mathrm{lat}_t,s_t,z_t,e_t),
\end{equation}
where $d_t$ denotes the day index, $(\mathrm{lon}_t,\mathrm{lat}_t)$ are geographic coordinates, $s_t$ denotes the speed-related signal, $z_t$ denotes navigation status and port-related information, and $e_t$ contains auxiliary contextual variables when available, the goal is to predict the next 30-day trajectory
\begin{equation}
\hat{\mathbf{Y}}=\{\hat{y}_t\}_{t=1}^{30},
\end{equation}
together with the final destination and arrival timing at the end of the horizon.

Unlike short-term coordinate extrapolation, month-level forecasting requires reasoning over voyage intent, route feasibility, maritime connectivity, and destination consistency. 
We formulate the task as structured prediction, where a policy LLM generates future trajectories and destination predictions from historical observations and maritime domain knowledge.

\subsection{Framework Overview}

The proposed model consists of two stages. First, raw AIS trajectories are transformed into structured textual prompts through trajectory preprocessing, segment-level feature extraction, trajectory semantic encoding, and prompt construction. In trajectory semantic encoding, location-consistent trajectory segments are converted into a semantic textual representation with a large-scale LLM, e.g., a 72B model, after movement, speed, turning, loitering, and transition features are extracted from the original AIS sequence. The resulting textual representation preserves route semantics and temporal context, and serves as the main input for downstream forecasting. Second, a smaller LLM policy model, e.g., a 4B--14B model, is directly optimized via verifier-guided reinforcement learning, without an intermediate supervised fine-tuning stage.

As shown in Fig.~\ref{fig:framework}, the proposed framework differs from conventional SFT$\rightarrow$RL pipelines. Instead of optimizing token-level imitation first, the proposed model aligns the policy directly with long-horizon forecasting objectives through a task-specific reward function that evaluates physical validity, process-level trajectory quality, and destination correctness.

\subsection{Trajectory Semantic Encoding and Prompt Construction}

In this stage, AIS records are converted into daily records,
\begin{equation}
q_t=(\text{date}_t,\text{lon}_t,\text{lat}_t,\text{speed}_t,\text{status}_t,\text{env}_t),
\end{equation}
where $\text{status}_t$ and $\text{env}_t$ denote navigation status and environmental context. For each $(\mathbf{X}_i,\mathbf{Y}_i)$ pair, location-consistent segments are extracted from $\mathbf{X}_i$ and encoded by a large-scale LLM into a semantic textual representation $\mathbf{H}_i$. The prompt-based RL dataset is represented as
\begin{equation}
\mathcal{D}_{\mathrm{RL}}=\{(\mathbf{P}_i,\mathbf{Y}_i)\}_{i=1}^{N},
\end{equation}
where $\mathbf{P}_i$ is constructed from $\mathbf{H}_i$ and task instructions, and $\mathbf{Y}_i$ is used as structured ground truth for verifier-based reward computation rather than supervised fine-tuning.

\subsection{Verifier-Guided Reinforcement Learning}

We optimize the forecasting policy using GRPO~\cite{shao2024deepseekmath} with a maritime verifiable reward. The reward pipeline evaluates output validity, physical feasibility, trajectory quality, and destination correctness through answer extraction, hard-constraint checking, process reward computation, and destination reward computation.

\subsubsection{Output Parsing and Hard Constraint Gating}

The model generates a response with \texttt{\textless think\textgreater} and \texttt{\textless answer\textgreater} segments, and the verifier parses the \texttt{\textless answer\textgreater} segment into a 30-day structured forecast. A missing \texttt{\textless answer\textgreater} tag receives zero reward, an incomplete or malformed answer may receive a small format-only reward, and severe physical violations terminate reward computation with a fixed penalty:
\begin{equation}
R_{\mathrm{hard}}=-0.5.
\end{equation}

The hard constraints include invalid output structure or trajectory length, malformed coordinates or day numbering, unrealistic daily displacement, and severe geographic infeasibility such as land crossing. To control unrealistic jumps, the maximum daily travel distance is bounded by
\begin{equation}
d_{\max}=v_{\max}\times 1.852 \times 24 \times k,
\end{equation}
where $1.852$ converts nautical miles to kilometers, $v_{\max}=25$ knots, and $k=1.2$, yielding approximately $1333.2$ km/day.

\subsubsection{Process Reward Model}

For valid outputs, a process reward is computed over Day 1--29:
\begin{equation}
r_t=\omega_{\mathrm{acc}}\,r_t^{\mathrm{acc}}+\omega_{\mathrm{corr}}\,r_t^{\mathrm{corr}},
\qquad t=1,\ldots,29,
\end{equation}
where $\omega_{\mathrm{acc}}=0.75$ and $\omega_{\mathrm{corr}}=0.25$. Here, $r_t^{\mathrm{acc}}$ measures geographic accuracy, while $r_t^{\mathrm{corr}}$ evaluates route-corridor feasibility based on layered reachability in the maritime graph.

Daily rewards are aggregated using exponential early weighting:
\begin{equation}
R_{\mathrm{traj}}=\sum_{t=1}^{29} w_t r_t,
\qquad
w_t=\frac{\delta^{\,t-1}}{\sum_{j=1}^{29}\delta^{\,j-1}},
\label{eq:rtraj_final}
\end{equation}
with $\delta=0.92$, assigning larger weights to earlier prediction days.

\subsubsection{Destination Reward}

The destination reward $R_{\mathrm{dest}}$ evaluates whether the forecast reaches the correct final destination region or port through hierarchical matching. The hierarchy includes exact or fuzzy location match, same port area, nearby distance thresholds, same sea body, same geographic region, hop-based reachability, same country, and unmatched destination. A Day 27–30 matching window is used to handle arrival-time uncertainty.

To provide graded reward signals, the destination module assigns partial credit across geographic levels and adds a Day-30 endpoint bonus for accurate arrivals. Exact or near-exact matches on Day 30 receive stronger bonuses, while weakly related destination matches receive reduced credit through a penalty multiplier.

\subsubsection{Curriculum Learning and Reward Fusion}

The trajectory and destination rewards are combined through a curriculum schedule:
\begin{equation}
R_{\mathrm{mix}}(\tau)=\alpha(\tau)R_{\mathrm{dest}}+\bigl(1-\alpha(\tau)\bigr)R_{\mathrm{traj}},
\end{equation}
where
\begin{equation}
\alpha(\tau)=\alpha_{\mathrm{init}}+(\alpha_{\mathrm{final}}-\alpha_{\mathrm{init}})
\min\!\left(1,\frac{\tau}{\tau_{\max}}\right),
\label{eq:curriculum_final}
\end{equation}
with $\alpha_{\mathrm{init}}=0.2$, $\alpha_{\mathrm{final}}=0.85$, and $\tau_{\max}=15000$.

The schedule shifts training from trajectory feasibility to destination correctness. The final reward adds a Day-30 endpoint bonus and is clipped for stability:
\begin{equation}
R_{\mathrm{final}} =
\operatorname{clip}\left(
R_{\mathrm{mix}} + B_{\mathrm{day30}},
-0.5, 1.0
\right).
\label{eq:final_reward}
\end{equation}

\subsection{Design Rationale}

The reward design addresses three key challenges in long-horizon maritime forecasting. First, hard-constraint gating suppresses physically implausible outputs and reduces reward hacking. Second, the PRM provides dense process-level supervision rather than relying only on the final endpoint. Third, hierarchical destination matching and curriculum fusion make reinforcement learning more stable by gradually shifting optimization from local trajectory feasibility to global destination accuracy. As a result, the proposed model treats long-horizon forecasting as a constrained structured forecasting problem over geography, maritime connectivity, and destination intent, rather than simple sequence extrapolation.


\section{Experiments}
\label{sec:experiments}

\subsection{Dataset Introduction}

\textbf{Data Collection and Processing:}  
The dataset is built from real-world AIS records of tanker vessels collected throughout 2022. The raw data include MMSI, GPS coordinates, speed over ground, and course over ground. To enrich spatial semantics, we further associate each AIS point with maritime water body labels using the World Seas IHO v3 shapefile.

The raw AIS streams are transformed into structured long-horizon trajectory samples. Vessel states are inferred using port proximity and speed heuristics, distinguishing in-port and underway behaviors. To handle irregular AIS reporting frequencies, all trajectories are resampled to hourly resolution by interpolation, and then converted into daily sequences by taking one record per day.

\textbf{Dataset Generation:}  
Training instances are created using a sliding-window strategy. Each sample contains a 60-day historical window and a 30-day prediction horizon, with windows moving forward by 30 days. Trajectories shorter than 90 days are excluded.

To provide higher-level semantic context, we augment each sample with semantic textual representations generated by a large-scale LLM, i.e., Qwen2.5-72B-Instruct. The semantic textual representations describe route evolution, navigational events, and destination-related semantics, and serve as the main prompt input for long-horizon forecasting.

The final dataset is organized in veRL format and split into 6,875 training and 779 test samples from 7,654 trajectory segments and 2,168 vessels, with MMSI information considered during the split. All methods are evaluated on the same 779-sample test set.

\subsection{Evaluation Metrics}

We evaluate model performance using both verifier-aligned reward metrics and interpretable geographic error metrics.


\paragraph{Destination Reward ($R_{\text{dest}}$)}
Endpoint quality is evaluated by hierarchical geographic matching, which assigns partial credit from exact match to weak regional match. The evaluation allows a Day 27–30 matching window for arrival time and assigns a Day-30 endpoint bonus to accurate arrivals.

\paragraph{Trajectory Reward ($R_{\text{traj}}$)}
The intermediate trajectory is evaluated over Day 1--29 using a process reward that combines daily position accuracy and route-corridor feasibility. The final trajectory reward uses exponentially decayed temporal weighting:
\begin{equation}
R_{\mathrm{traj}}=\sum_{t=1}^{29} w_t r_t,\qquad
w_t=\frac{\delta^{t-1}}{\sum_{j=1}^{29}\delta^{j-1}},
\end{equation}
where $\delta=0.92$ assigns larger weight to early prediction days.

\paragraph{Combined Score}
The final reward-based score is defined as
\begin{equation}
\mathrm{Score}=\operatorname{clip}\left(
\alpha R_{\mathrm{dest}}+(1-\alpha)R_{\mathrm{traj}}+B_{\mathrm{day30}},
-0.5, 1.0
\right),
\end{equation}
where $\alpha=0.85$ at evaluation time. $B_{\mathrm{day30}}$ is an additional endpoint bonus computed on Day 30, while $R_{\mathrm{dest}}$ uses a Day 27--30 matching window to handle arrival-time uncertainty.

\subsection{Training Configuration}

The proposed model is trained with GRPO using 4 rollouts per prompt, learning rate $1\times10^{-6}$ with 10\% warmup, batch size 112, PPO mini-batch size 56, micro-batch size 7 per GPU, and KL coefficient 0.001. The maximum prompt and response lengths are 13,000 and 8,000 tokens. Rollout generation uses temperature 0.8, top-$k=40$, top-$p=0.9$, and repetition penalty 1.25. The curriculum weight $\alpha$ increases from 0.2 to 0.85 over 15,000 steps, with $\omega_{\text{acc}}=0.75$ and $\omega_{\text{corr}}=0.25$. Training uses 8 NVIDIA A100 80GB GPUs with FSDP, gradient checkpointing, and vLLM rollout generation under tensor parallelism of one.

\section{Results and Discussion}

\subsection{Zero-Shot LLM and Spatio-Temporal LLM Baselines}

\begin{table}[t]
\centering
\caption{Performance of zero-shot and spatio-temporal LLM baselines on 30-day maritime trajectory forecasting.}
\label{tab:llm_baselines}
\resizebox{\columnwidth}{!}{%
\begin{tabular}{l r r c}
\toprule
\textbf{Model} & \textbf{Traj.}$\downarrow$ & \textbf{Endpoint}$\downarrow$ & \textbf{Parse}$\uparrow$ \\
 & \textbf{Err.} (km) & \textbf{Err.} (km) & (\%) \\
\midrule
\multicolumn{4}{l}{\textit{Zero-shot LLM Baselines}} \\
Qwen3-4B & 3,177.9 & 3,860.7 & 97.9 \\
Qwen3-4B-Thinking-2507 & 3,138.6 & 3,906.6 & 62.8 \\
Qwen3-8B & 3,084.9 & 4,010.7 & 96.7 \\
Llama-3.1-Tulu-3-8B-SFT & 3,563.9 & 4,073.3 & 93.2 \\
\midrule
\multicolumn{4}{l}{\textit{LLM-based Spatio-temporal Baselines}} \\
ST-LLM & 2,014.1 & 2,909.5 & \textbf{100.0} \\
ST-LLM+ & \textbf{1,956.1} & \textbf{2,820.9} & \textbf{100.0} \\
\bottomrule
\end{tabular}
}
\vspace{1mm}
\begin{flushleft}
\end{flushleft}
\end{table}

Table~\ref{tab:llm_baselines} reports the performance of LLM-based baselines on maritime trajectory forecasting using the same evaluation metrics as the deep learning baselines. We consider two categories of LLM-based methods. The first category is zero-shot LLM baselines, including Qwen3-4B, Qwen3-8B, Llama-3.1-Tulu-3-8B-SFT~\cite{tulu3_sft,samragh2025future}, and Qwen3-4B-Thinking-2507. These models are prompted with 60 days of historical vessel information, including location, status, speed, wind, and significant wave height, and generate structured 30-day forecasts without task-specific training. A prediction is considered valid only if it can be parsed into a complete 30-day schedule.

The second category is trained spatio-temporal LLM baselines, including ST-LLM~\cite{liu2024stllm} and ST-LLM+~\cite{liu2025stllmplus}. We implement ST-LLM and ST-LLM+ using the official code repositories and adapt the input format to the maritime dataset by treating latitude and longitude as two spatio-temporal channels. For ST-LLM+, a compact latitude–longitude channel graph is used to model the dependency between the two coordinate dimensions. As adapted baselines, ST-LLM and ST-LLM+ are trained on the maritime dataset and evaluated by comparing predicted latitude–longitude sequences with ground-truth trajectories. As shown in Table~\ref{tab:llm_baselines}, zero-shot LLMs achieve high parse rates in most cases but still suffer from large trajectory and endpoint errors. In contrast, the trained spatio-temporal LLM baselines substantially reduce both errors, with ST-LLM+ achieving the best baseline performance.

\subsection{Comparison Among RLVR-Trained LLM Variants}

Tables~\ref{tab:training} and \ref{tab:evaluation} summarize the training and test performance of RLVR-trained LLM variants. Here, $R_{\text{dest}}$, $R_{\text{traj}}$, and RLVR score are verifiable reward metrics, while endpoint error and trajectory error are geographic distance metrics. Overall, Gemma3-4B performs best, achieving the highest reward-based metrics and the lowest distance errors on the test set. Qwen3-4B-Instruct ranks second, while the Qwen3 base models are less competitive. The larger Qwen3-8B and Qwen3-14B do not outperform the smaller 4B variants, suggesting that model scale alone does not guarantee better performance in this reward-gated structured forecasting task.


\begin{table}[htbp]
\centering
\caption{Training performance of RLVR-trained LLM variants.}
\label{tab:training}
\begin{tabular}{lccc}
\toprule
\textbf{Model} & $\mathbf{R_{\text{dest}}}\uparrow$ & $\mathbf{R_{\text{traj}}}\uparrow$ & \textbf{RLVR Score}$\uparrow$ \\
\midrule
Gemma3-4B         & \textbf{0.50} & \textbf{0.80} & \textbf{0.53} \\
Qwen3-4B-Instruct & 0.32 & 0.75 & 0.41 \\
Qwen3-4B          & 0.28 & 0.70 & 0.38 \\
Qwen3-8B          & 0.29 & 0.67 & 0.35 \\
Qwen3-14B         & 0.13 & 0.60 & 0.22 \\
\bottomrule
\end{tabular}
\end{table}



\begin{table}[htbp]
\raggedright
\caption{Evaluation performance of RLVR-trained LLM variants.}
\label{tab:evaluation}
\resizebox{\columnwidth}{!}{%
\begin{tabular}{lccccc}
\toprule
\textbf{Model} & $\mathbf{R_{\text{dest}}}\uparrow$ & $\mathbf{R_{\text{traj}}}\uparrow$ & \textbf{RLVR}$\uparrow$ & \textbf{Traj.}$\downarrow$ & \textbf{Endpoint}$\downarrow$ \\
 &  &  & \textbf{Score} & \textbf{Err.} (km) & \textbf{Err.} (km) \\
\midrule
Gemma3-4B         & \textbf{0.50} & \textbf{0.80} & \textbf{0.49} & \textbf{100 $\pm$ 85} & \textbf{465 $\pm$ 380} \\
Qwen3-4B-Instruct & 0.32 & 0.75 & 0.39 & 130 $\pm$ 95 & 520 $\pm$ 420 \\
Qwen3-4B          & 0.28 & 0.70 & 0.38 & 150 $\pm$ 110 & 580 $\pm$ 450 \\
Qwen3-8B          & 0.29 & 0.67 & 0.35 & 180 $\pm$ 130 & 605 $\pm$ 470 \\
Qwen3-14B         & 0.13 & 0.60 & 0.22 & 220 $\pm$ 160 & 780 $\pm$ 520 \\
\bottomrule
\end{tabular}%
}
\end{table}

\begin{figure}[t]
    \centering
    \includegraphics[width=\columnwidth]{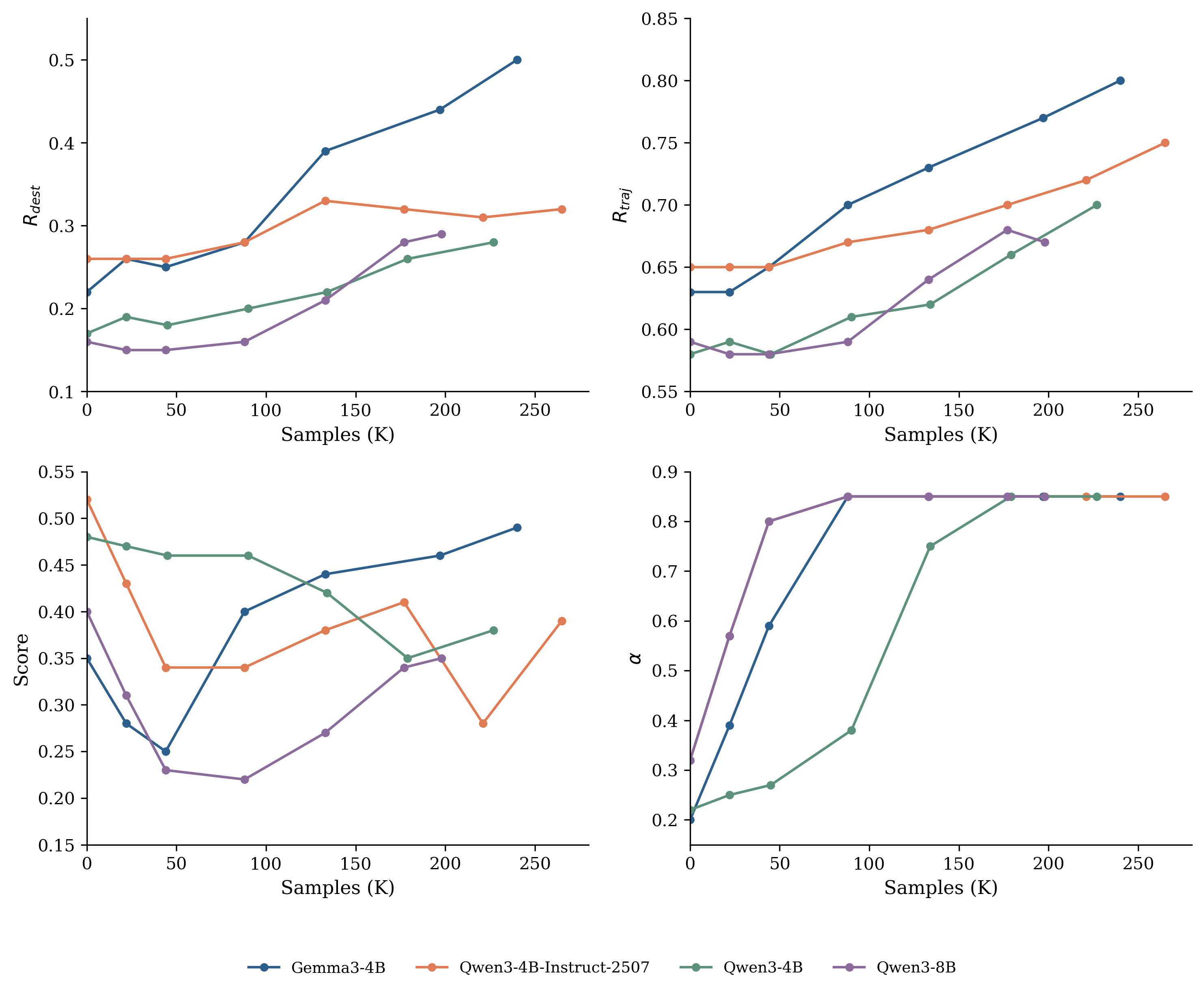}
    \caption{Training dynamics of four RLVR-trained LLM variants under verifier-guided GRPO. The subplots show destination reward $R_{\text{dest}}$, trajectory reward $R_{\text{traj}}$, RLVR score, and curriculum parameter $\alpha$.}
    \label{fig:training_curves}
\end{figure}


Figure~\ref{fig:training_curves} shows that all models improve in $R_{\text{traj}}$, while the main difference lies in $R_{\text{dest}}$. Gemma3-4B achieves the strongest gains in both rewards and obtains the highest final RLVR score. Qwen3-4B-Instruct improves steadily but remains behind Gemma3-4B, while Qwen3-4B and Qwen3-8B converge to weaker solutions. The score curves also reflect the curriculum schedule: as $\alpha$ increases, training places more weight on destination accuracy, and models with stronger $R_{\text{dest}}$ improvement achieve better final performance.

\begin{figure}[t]
    \centering
    \includegraphics[width=\columnwidth]{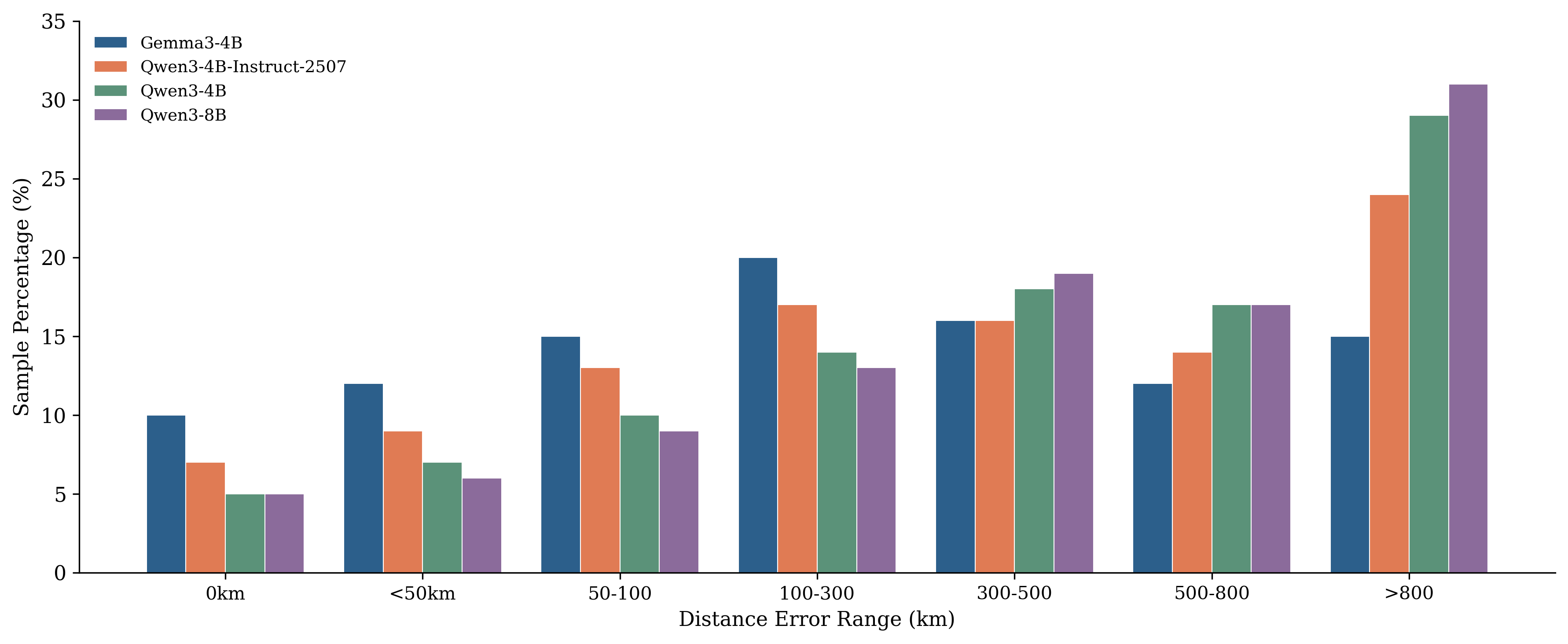}
    \caption{Distribution of destination endpoint error ranges on the test set for different RLVR-trained LLM variants. The y-axis shows the percentage of samples in each error interval.}
    \label{fig:error_distribution}
\end{figure}


Figure~\ref{fig:error_distribution} complements Table~\ref{tab:evaluation} by showing the endpoint-error distribution. Gemma3-4B has a larger proportion of samples in the low-error ranges and a smaller proportion in the $>800$ km range, consistent with the lowest average endpoint error in Table~\ref{tab:evaluation}. Qwen3-4B-Instruct shows a similar but less concentrated pattern. In contrast, Qwen3-4B and Qwen3-8B have heavier high-error tails, indicating that these variants produce large destination errors more frequently.


Overall, the RLVR results indicate that the best performance is achieved when model capacity, reward structure, and optimization stability are well matched. The stronger performance of the 4B variants should therefore be interpreted as a task-specific capacity match rather than a general scaling conclusion. The maritime RLVR task is highly structured and reward-gated, requiring valid 30-day trajectory forecasts and destination predictions under a strict output schema. Under the fixed optimization budget, 4B models are sufficiently expressive to capture semantic maritime patterns and easier to optimize toward reward-compatible outputs. Larger LLM variants have stronger general capacity in principle, but additional capacity does not necessarily improve final reward under strict parsing, hard-constraint gating, and curriculum-based optimization. This result suggests that, for the current RLVR setting, training stability and reward compatibility are more critical than model scale alone.

\subsection{Comparison with Deep Learning Baselines}



We compare the proposed model with representative deep learning baselines, including LSTM, Informer, Autoformer, FEDformer, PatchTST, TimesNet, and iTransformer. All baselines take 60-day historical observations as input and predict the next 30-day trajectory, and are trained for 50 epochs with Adam and a learning rate of $10^{-4}$.

Table~\ref{tab:dl_baselines} shows that LSTM achieves the best performance among the deep learning baselines, with the lowest trajectory and endpoint errors and the highest verifier-based score. Endpoint errors are larger than trajectory errors for all models, indicating error accumulation toward the final prediction day. Autoformer, TimesNet, and iTransformer obtain similar intermediate results, while PatchTST performs the worst on the reward-based metrics.


\begin{table}[t]
\centering
\caption{Performance of deep learning baselines on 30-day maritime trajectory forecasting.}
\label{tab:dl_baselines}
\resizebox{\columnwidth}{!}{%
\begin{tabular}{l c c c c c}
\toprule
\textbf{Model} & \textbf{Traj.}$\downarrow$ & \textbf{Endpoint}$\downarrow$ & $\mathbf{R_{\text{dest}}}\uparrow$ & $\mathbf{R_{\text{traj}}}\uparrow$ & \textbf{RLVR Score}$\uparrow$ \\
 & \textbf{Err.} (km) & \textbf{Err.} (km) &  &  &  \\
\midrule
\multicolumn{6}{l}{\textit{Recurrent Neural Networks}} \\
LSTM & \textbf{1,934} & \textbf{2,918} & \textbf{0.207} & \textbf{0.375} & \textbf{0.258} \\
\midrule
\multicolumn{6}{l}{\textit{Transformer-Based Architectures}} \\
Informer & 2,875 & 3,150 & 0.161 & 0.358 & 0.212 \\
Autoformer & 2,841 & 3,109 & 0.188 & 0.352 & 0.235 \\
FEDformer & 2,954 & 3,284 & 0.170 & 0.352 & 0.221 \\
\midrule
\multicolumn{6}{l}{\textit{Recent Advanced Models}} \\
PatchTST & 3,049 & 3,384 & 0.146 & 0.305 & 0.189 \\
TimesNet & 2,897 & 3,097 & 0.188 & 0.352 & 0.238 \\
iTransformer & 2,907 & 3,149 & 0.190 & 0.349 & 0.238 \\
\bottomrule
\end{tabular}
}
\end{table}



The comparison reflects the different learning mechanisms of the RLVR-trained LLM and the deep learning baselines. The proposed task is designed to adapt pretrained LLM capabilities to maritime forecasting through trajectory textualization, structured prompting, and verifier-guided optimization, and thus the model learns from semantic voyage context rather than only coordinate sequences. By contrast, the deep learning baselines perform direct coordinate regression from 60-day latitude–longitude histories to 30-day future positions. Under this setting, LSTM is well matched to the available data scale and input form, as several thousand training samples are sufficient for learning local vessel-position continuity. Transformer-style spatio-temporal models usually benefit more from larger datasets and richer structured inputs, such as port tokens, route graphs, vessel attributes, traffic patterns, and maritime topology. Therefore, the LSTM result reflects a strong coordinate-regression baseline under the current DL pipeline, rather than a general superiority over modern spatio-temporal architectures.


\section{Conclusion}
\label{sec:conclusion}
This paper investigates joint long-horizon vessel trajectory and destination forecasting with reasoning-oriented large language models. The study develops a maritime LLM post-training framework with verifiable reward, in which AIS trajectories are converted into semantic textual representations for RL prompt construction. The proposed RLVR framework aligns the policy model with maritime forecasting objectives by enforcing physical validity, providing process-level trajectory supervision, and rewarding destination correctness through curriculum learning.

The results show that RLVR training substantially improves over zero-shot LLMs and widely used deep learning baselines, especially on destination-related metrics. 
Among the evaluated RLVR-trained LLM variants, 4B models achieve the best results, indicating a task-specific capacity match rather than a general scaling trend.
Under strict parsing, hard-constraint gating, and a fixed optimization budget, 4B models are expressive enough to capture semantic maritime patterns and easier to optimize toward reward-compatible outputs than larger LLMs. For deep learning baselines, LSTM outperforms several widely used forecasting models under the current setting. Since the task is designed for LLM post-training with limited fine-tuning data, the DL baselines mainly perform direct coordinate regression from latitude–longitude sequences, where a compact recurrent model can effectively learn local vessel-position continuity. Transformer-style spatio-temporal models typically require larger datasets and richer structured inputs to fully exploit their capacity.

Several limitations remain. The benchmark is restricted to tanker vessels from one year, and the verifiable reward still contains rule-based components. Future work will extend the data coverage, incorporate richer maritime context, and develop more adaptive verifiable rewards for multimodal long-horizon forecasting and decision support.

\section*{Acknowledgment}

This work was supported in part by the Maritime AI Research Programme phase 2 (project “Maritime Artificial Intelligence (AI) Research for Shipping Disruption Evaluation, Launch Boat Optimisation and Digital Testing of Vessel Predictive Maintenance” with grant number SMI-2025-MTP-02 funded by the Singapore Maritime Institute).

\bibliographystyle{IEEEtran}
\bibliography{root}

\end{document}